\documentclass[letterpaper, 10 pt, journal, twoside]{IEEEtran}
\usepackage{graphicx} 
\usepackage{amsmath,amssymb}
\usepackage{nth}
\usepackage{color} %red, green, blue, yellow, cyan, magenta, black, white
\definecolor{mygreen}{RGB}{28,172,0} % color values Red, Green, Blue
\definecolor{mylilas}{RGB}{170,55,241}
\usepackage{makecell}
\usepackage[utf8]{inputenc}
\usepackage{glossaries}
\usepackage{subcaption}
\usepackage[export]{adjustbox}
\usepackage{multirow}
\usepackage{svg}
\usepackage{wrapfig}
\usepackage[textsize=scriptsize]{todonotes}

\usepackage[linesnumbered,ruled,vlined]{algorithm2e}

\SetCommentSty{mycommfont}
\usepackage{xcolor}

\def\HiLi{\leavevmode\rlap{\hbox to \hsize{\color{yellow!50}\leaders\hrule height .8\baselineskip depth .5ex\hfill}}}

\DeclareMathOperator*{\argmin}{arg\,min}

\usepackage{lipsum}
\usepackage{float}
\makeatletter
\def\BState{\State\hskip-\ALG@thistlm}
\makeatother
\newcommand{\norm}[1] {||#1||}
\usepackage{tabularx,ragged2e,booktabs,caption}

\newcolumntype{C}[1]{>{\Centering}m{#1}}

\newcommand{\normal}{\mathcal{N}}

\usepackage{flexisym}
\newcommand{\BEAS}{\begin{eqnarray*}}
\newcommand{\EEAS}{\end{eqnarray*}}
\newcommand{\BEA}{\begin{eqnarray}}
\newcommand{\EEA}{\end{eqnarray}}
\newcommand{\BEQ}{\begin{equation}}
\newcommand{\EEQ}{\end{equation}}
\newcommand{\BIT}{\begin{itemize}}
\newcommand{\EIT}{\end{itemize}}

\newcommand{\st}{\text{s.t.}}

\newcommand{\reals}{{\mbox{\textbf{R}}}}

\newcommand{\diag}{\mathop{\textbf{diag}}}

\newcommand{\Expect}{\mathop{\mathbb{E}}}
\newcommand{\var}{\mathop{\textbf{var}}} % variance
   {\begin{list}{}{%
    \setlength{\rightmargin}{0\linewidth}%
    \setlength{\leftmargin}{.05\linewidth}}%
    \sffamily\small
    \item[]{\setlength{\parskip}{0ex}\hrulefill\par%
    \nopagebreak{}}}%
   {{\setlength{\parskip}{-1ex}\nopagebreak\par\hrulefill} \end{list}}

\graphicspath{{media/}}
\usepackage{mydefs}

\usepackage{hyperref}

\ifCLASSINFOpdf
\else
\fi
\begin{document}
\title{\LARGE \bf
\controllerabv{}: A Controller for Escaping Traps in Novel Environments}

% \author{Michael~Shell,~\IEEEmembership{Member,~IEEE,}
%         John~Doe,~\IEEEmembership{Fellow,~OSA,}
%         and~Jane~Doe,~\IEEEmembership{Life~Fellow,~IEEE}% <-this % stops a space
% \thanks{M. Shell was with the Department
% of Electrical and Computer Engineering, Georgia Institute of Technology, Atlanta,
% GA, 30332 USA e-mail: (see http://www.michaelshell.org/contact.html).}% <-this % stops a space
% \thanks{J. Doe and J. Doe are with Anonymous University.}% <-this % stops a space
% \thanks{Manuscript received April 19, 2005; revised August 26, 2015.}}
\author{Sheng Zhong$^{1}$, Zhenyuan Zhang$^{1}$, Nima Fazeli$^{1}$, and Dmitry Berenson$^{1}$%
\thanks{Manuscript received: October, 15, 2020; Revised January, 6, 2021; Accepted February, 1, 2021.}%Use only for final RAL version
\thanks{This paper was recommended for publication by Editor Dana Kulic upon evaluation of the Associate Editor and Reviewers' comments.
This work was supported in part by NSF grant IIS-1750489.} %Use only for final RAL version
\thanks{$^{1}$Robotics Institute, University of Michigan, MI 48109, USA
        {\tt\footnotesize \{zhsh, cryscan, nfz, dmitryb\}@umich.edu}}%
\thanks{Digital Object Identifier (DOI): see top of this page.}
\thanks{For code, see \href{https://github.com/UM-ARM-Lab/tampc}{\tt\small https://github.com/UM-ARM-Lab/tampc}}%
}

\markboth{IEEE Robotics and Automation Letters. Preprint Version. Accepted February, 2021}
{Zhong \MakeLowercase{\textit{et al.}}: TAMPC: A Controller for Escaping Traps in Novel Environments} 




\maketitle

\begin{abstract}
We propose an approach to online model adaptation and control in the challenging case of hybrid and discontinuous dynamics where actions may lead to difficult-to-escape ``trap'' states, under a given controller. We first learn dynamics for a system without traps from a randomly collected training set
(since we do not know what traps will be encountered online). These ``nominal'' dynamics allow us to perform tasks 
in scenarios where the dynamics matches the training data,
but when unexpected traps arise in execution, we must find a way to adapt our dynamics and control strategy and continue attempting the task. Our approach, \controllername{} (\controllerabv), is a two-level hierarchical control algorithm that reasons about traps and non-nominal dynamics to decide between goal-seeking and recovery policies.
An important requirement of our method is the ability to recognize nominal dynamics even when we encounter data that is out-of-distribution w.r.t the training data. We achieve this by learning a representation for dynamics that exploits invariance in the nominal environment, thus allowing better generalization. We evaluate our method on simulated planar pushing and peg-in-hole as well as real robot peg-in-hole problems against adaptive control, reinforcement learning, trap-handling baselines, where traps arise due to unexpected obstacles that we only observe through contact. Our results show that our method outperforms the baselines on difficult tasks, and is comparable to prior trap-handling methods on easier tasks.
\end{abstract}

% \begin{IEEEkeywords}
% IEEE, IEEEtran, journal, \LaTeX, paper, template.
% \end{IEEEkeywords}

% \fi
\IEEEpeerreviewmaketitle

\section{INTRODUCTION}
\label{sec:introduction}

\IEEEPARstart{I}{n} this paper, we study the problem of controlling robots in environments with unforeseen \textbf{traps}. Informally, traps are states in which the robot's controller fails to % cannot\todo{Brad: "traps are states in which the robot's controller fails to..."} 
make progress towards its goal and gets ``stuck". Traps are common in robotics and can arise due to many factors including geometric constraints imposed by obstacles, frictional locking effects, and nonholonomic dynamics leading to dropped degrees of freedom~\cite{borenstein2005omnitread},~\cite{fantoni2002non},~\cite{koditschek2004mechanical}.  
%
% For example, frictional jamming induces trap dynamics for object manipulation in clutter. We consider a simplified form of this problem in planar pushing with walls.
% \jznote{Additionally, we consider peg-in-hole problems with unmodeled obstacles near the goal.}
In this paper, we consider instances of trap dynamics in planar pushing with walls and peg-in-hole with unmodeled obstructions to the goal.
% In peg-in-hole problems, traps could result from imperfections on the surface containing the hole. The strategy of sliding the peg along the surface could catch on an imperfection, leading to the robot being stuck.
% manipulator arm gets some of its upper joints caught in an unmodeled obstacle -- yes, when does this happen though in the real world?
% grasping in cluttered environments with obstacles/other objects behind what we're targeting?

Developing generalizable algorithms that rapidly adapt to handle the wide variety of traps encountered by robots is important to their deployment in the real-world. Two central challenges in online adaptation to environments with traps are the data-efficiency requirements and the lack of progress towards the goal for actions inside of traps. 
In this paper, our
key insight is that we can address these challenges by explicitly reasoning over different dynamic modes, in particular traps, together with contingent recovery policies, organized as a hierarchical controller.
We introduce an online modeling and controls method that balances naive optimism and pessimism when encountering novel dynamics. Our method learns a dynamics representation that infers underlying invariances %, such as translational invariance in planar problems, and} \nfnote{this translational invariance thing is not reading well -- it's also not necessarily true}
% an invariant\todo{invariant to what?} dynamics representation and 
and exploits it when possible (optimism) while treading carefully to escape and avoid potential traps in non-nominal dynamics (pessimism). Specifically, we:
\begin{enumerate}
    \item % Introduce a novel representation architecture\todo{D: Should this be "representation-learning"?} for generalizing dynamics %, show how to learn it offline from nominal data, 
    %and show how it allows our method to achieve good performance on out-of-distribution data;
    Introduce a novel representation learning approach that effectively generalizes dynamics and show its efficacy for task execution on out-of-distribution data when used with our proposed controller;
    \item Introduce \controllername{} (TAMPC), a novel control algorithm that reasons about non-nominal dynamics and traps to reach goals in novel environments with traps;
    \item Evaluate our method on real robot and simulated peg-in-hole, and simulated planar pushing tasks with traps where adaptive control and reinforcement learning baselines achieve 0\% success rate. These include difficult tasks where trap-handling baselines achieve less than 50\% success, while our method achieves at least 60\% success on all tasks. 
\end{enumerate}

We show that state-of-the-art techniques~\cite{fu2016one},~\cite{haarnoja2018soft}, while capable of adapting to novel dynamics, are insufficient for escaping traps that our approach handles by their explicit consideration. Additionally, our method performs well on tasks that prior trap-handling methods struggle on.
% Our approach addresses limitations in state-of-the-art techniques~\cite{fu2016one},~\cite{haarnoja2018soft} that cannot be expected to perform well in these scenarios because they have little to no contingencies for handling traps -- action sequences that escape traps incur high short-term costs and are much less likely to be discovered. 

% In this paper, our key insight is to explicitly reason over different dynamic modes, in particular traps, balancing optimistic exploitation of known dynamics towards the goal and policies intended to recover from traps. 
% Our method learns an invariant dynamics representation and exploits it when possible (optimism) while treading carefully to escape and avoid potential traps in new environments (pessimism). Specifically, we:
\begin{figure}[t]
    \centering
    \includegraphics[width=0.49\textwidth]{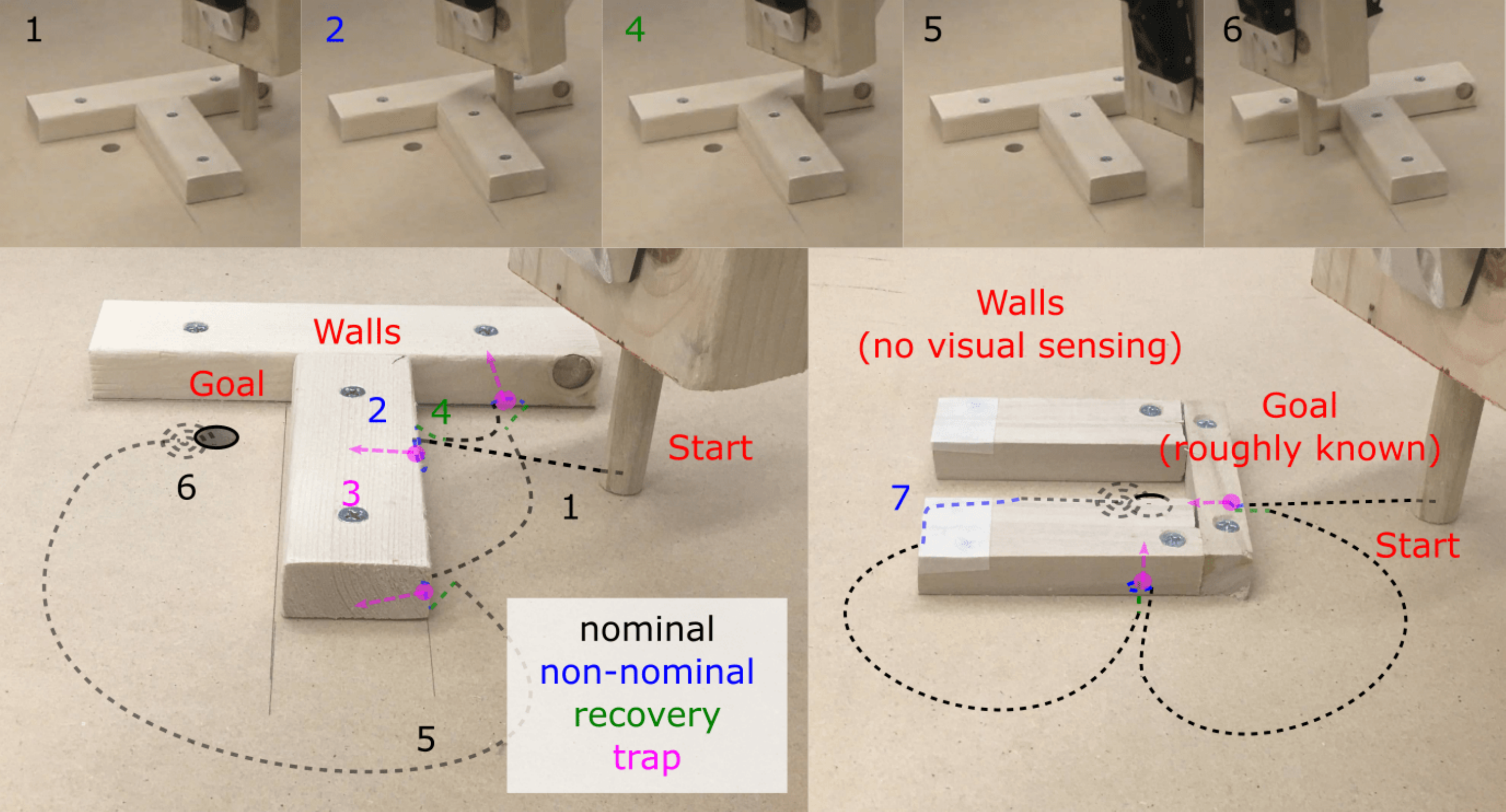}
    \caption{\controllerabv{} on peg-in-hole tasks with obstacles near the goal. The robot has no visual sensing (cannot anticipate walls) and has not encountered walls in the training data. Path segments show (1) the initial direct trajectory to goal, (2) detecting non-nominal dynamics from reaction forces and exploration of it by sliding along the wall, (3) detecting a trap due to the inability to make progress using non-nominal dynamics, (4) recovery to nominal dynamics, (5) going around seen traps to goal, (6) spiraling to find the precise location of the hole, and (7) sliding against the wall (non-nominal) towards the goal.}
    \label{fig:real peg tasks}
    \vspace{4pt}
\end{figure}

\section{PROBLEM STATEMENT}
Let $\x \in \xall$ and $\us \in \usall$ denote the $\nx$ dimensional state and $\nus$ dimensional control.
Under test conditions the system follows novel dynamics 
% $\newdynamics: \xall \times \usall \rightarrow \tangent \xall$. \todo{What does the tangent mean?}
$\dx = \newdynamics(\x, \us)$.
The objective of our method is to reach a goal $\x \in \goal \subset \xall$ as quickly as possible:
\begin{equation}\label{eq:optimal objective}
    \begin{array}{cl}
    \argmin\limits_{\us_0,...,\us_{\nsteps-1}}&\nsteps\\ 
    \st&\x_{t+1} = \x_t + \newdynamics(\x_t, \us_t), \ t=0,...,\nsteps-1\\
    &\x_\nsteps \in \goal\\
    \end{array}
\end{equation}

This problem is difficult because the novel dynamics $\newdynamics$ are not known.
Instead, we assume we have access to a dataset of sampled transitions with random actions, which can be used to learn an approximate dynamics model $\learnednom$ of the system under \textit{nominal} dynamics $\nomdynamics$ where:
\begin{align}
    \newdynamics(\x,\us) = \nomdynamics(\x,\us) + \errdynamics(\x,\us)
\end{align}
We assume the error dynamics $\errdynamics$ are relatively small (w.r.t. nominal) except for non-nominal regions $\xnonnom \subset \xall$ for which: 
\begin{equation}
    |\errdynamics(\x,\us)| \sim |\nomdynamics(\x,\us)|, \quad \quad \forall \x \in \xnonnom, \; \; \exists \; \us \in \usall\label{eq:nonnominal dynamics}
\end{equation}

In nominal regions where $\learnednom \approx \nomdynamics \approx \newdynamics$ a Model Predictive Controller (MPC) can provide high quality solutions without direct access to $\newdynamics$. Using a specified cost function $\costfunc: \xall \times \usall \rightarrow [0,\infty)$, following the MPC policy $\us = \controller{}(\x, \learnednom, \costfunc)$ creates the dynamical system:
\begin{equation}
    \dx = \newdynamics(\x,\controller(\x,\learnednom,\costfunc))
    \label{eq:closed sys}
\end{equation}

This dynamical system may have attractors~\cite{milnor1985concept}, which are subsets $\attractor \subseteq \xall$ where:
% \begin{align}
%     \x_{t_0} \in \attractor \implies \x_t \in \attractor \: \: \: \: \: \forall t \ge t_0
% \end{align}
\begin{itemize}
    \item $\x_{t_0} \in \attractor \implies \x_t \in \attractor  \: \: \: \: \: \forall t \ge t_0$
    \item $\attractor$ has a basin of attraction \\$B(\attractor) \subseteq \xall = \{\x \ | \ \x_0=\x,\ \exists \ t \ge 0,\ \x_t \in \attractor\}$ 
    \item $\attractor$ has no non-empty subset with the first two properties
\end{itemize}
% Before defining traps, we define an attractor~\cite{milnor1985concept},~\cite{dupuis2015escaping} of the dynamical system as any subset $\attractor \subseteq \xall$ that: }
% \jznote{
% \begin{itemize}
%     \item is forward invariant: $\xs_{t_0} \in \attractor \implies \xs_t \in \attractor \ \forall t \ge t_0$
%     \item has a basin of attraction \\$B(\attractor) \subseteq \xall = \{\x \ | \ \xs_0=\x,\ \exists \ t \ge 0,\ \xs_t \in \attractor\}$ 
%     \item has no non-empty subset with the first two properties
% \end{itemize}
%  We define a \textbf{trap} as an attractor $\attractor$ such that $\attractor \cap \goal = \emptyset$, and the trap set $\trap \subseteq \xall$ as the union of all traps.
%  }
We define a \textit{trap} as an attractor $\attractor$ such that $\attractor \cap \goal = \emptyset$, and the \textit{trap set} $\trap \subseteq \xall$ as the union of all traps. % A \textit{trap basin} is defined as $B(\attractor) \subseteq \xall = \{\x \ | \ \x_0=\x,\ \exists \ t \ge 0,\ \x_t \in \attractor\}$. 
Escaping a trap requires altering the dynamical system (Eq. \ref{eq:closed sys}) by altering $\learnednom$, $\costfunc$, or the $\controller{}$ function. 
Traps can be avoided in nominal regions with a sufficiently-powerful and long-horizon MPC, and a sufficiently-accurate dynamics approximation $\learnednom$.
This work addresses detecting, recovering from, and avoiding traps caused by non-nominal dynamics.
To aid in trap detection, recovery, and avoidance we assume a state distance function $\xdist : \xall \times \xall \rightarrow [0, \infty)$ and a control similarity function $\usim : \usall \times \usall \rightarrow [0,1]$ are given.

\section{RELATED WORK}
\label{sec:related work}
In this section, we review related work to the two main components of this paper: handling traps and generalizing models to out-of-distribution (OOD) dynamics.

\textbf{Handling Traps:} Traps can arise due to many factors including nonholonomic dynamics, frictional locking, and geometric constraints~\cite{borenstein2005omnitread},~\cite{fantoni2002non},~\cite{koditschek2004mechanical}. 
In particular, they can occur when the environment is partially unknown, as in the case of online path planning.
% In the control literature, viability control \cite{panagou2013viability} can be applied to nonholonomic systems in the case where the dynamics of entering traps (leaving the viability set) is known. 
% While they enforce staying inside the viability set as a hard constraint, our method can be interpreted as online learning of the non-viable set with a policy for returning to the viable set. 
%with a cost to penalize returning to the known non-viable set so far.

Traps have been considered in methods based on Artificial Potential Fields (APF)~\cite{khatib1986real},~\cite{lee2003artificial}. % ,~\cite{bounini2017modified}. 
Traps are local minima in the potential field under the policy of following the gradient of the field. In the case where the environment is only partially known \textit{a priori}, local minima escape (LME) methods such as~\cite{lee2003artificial},~\cite{fedele2018obstacles} can be used to first detect then escape local minima. 
% can be performed by first detecting the local minima, then placing a virtual obstacle, typically in the form of a positive ball in the potential, to drive the robot away and avoid revisits~\cite{lee2003artificial},~\cite{chengqing2000virtual}. 
Our method uses similar ideas to virtual obstacles (VO)~\cite{lee2003artificial} for detecting traps and avoiding revisits while addressing weaknesses common to APF-LME methods. Specifically, we are able to handle traps near goals by 
associating actions with trap states (using $\usim$) to penalize similar actions to the one previously entering the trap while near it, rather than penalize being near the trap altogether.
% conditioning the seen traps on the action taken using the control similarity function $\usim$.
We also avoid blocking off paths to the goal with virtual obstacles by shrinking their effect over time. Lastly, having an explicit recovery policy and using a controller that plans ahead lets us more efficiently escape traps with ``deep" basins (many actions are required to leave the basin). 
% that APF methods must iteratively ``fill up". 
We compare against two APF-LME methods in our experiments.

Another way to handle traps is with exploration, such as through intrinsic curiosity (based on model predictability)~\cite{pathak2017curiosity},~\cite{burda2018exploration}, state counting~\cite{bellemare2016unifying}, or stochastic policies~\cite{osband2016deep}. However, trap dynamics can be difficult to escape from and can require returning to dynamics the model predicts well (so receives little exploration reward). We show in an ablation test how random actions are insufficient for escaping traps in some tasks we consider. Similar to~\cite{ecoffet2019go}, we remember interesting past states. While they design domain-specific state interest scores, we effectively allow for online adaptation of the state score based on how much movement the induced policy generates while inside a trap. We use this score to direct our recovery policy. % and attempt to recover to them \jzc{to escape traps.} They require \jzc{simulator resets} and design domain-specific state interest scores. We do not require resetting, and effectively allow for online adaptation of the state score based on how much movement the induced policy generates while inside a trap.

Adapting to trap dynamics is another possible approach. Actor-critic methods have been used to control nonlinear systems with unknown dynamics online~\cite{dierks2012online}, % ~\cite{bhasin2013novel}
 and we evaluate this approach on our tasks.
% However, they tend to do poorly in discontinuous dynamics and are sample inefficient compared to our method as we show in experimentation. 
Another approach %to handle novel environments online
is with locally-fitted models %~\cite{levine2014learning},
which~\cite{fu2016one} showed could be mixed with a global model and used in MPC.
Similar to this approach, our method adapts a nominal model to local dynamics; however, we do not always exploit the dynamics to reach the goal. 
%In experimentation, we comapre against these methods and show that just adapting the model achieves poor performance on tasks involving traps.

\textbf{Generalizing models to OOD Dynamics:}
One goal of our method is to generalize the nominal dynamics to OOD novel environments. A popular approach for doing this is explicitly learning to be robust to expected variations across training and test environments. This includes methods such as meta-learning~\cite{finn2017model},~\cite{li2018learning}, domain randomization~\cite{tobin2017domain},~\cite{pan2010domain}, Sim-to-real~\cite{peng2018sim},
%~\cite{james2019sim}, 
and other transfer learning~\cite{zhang2018decoupling} methods. These methods are unsuitable for this problem because our training data contains only nominal dynamics, whereas they need a diverse set of non-nominal dynamics. Instead, we
% A specific way of doing this is to 
learn a robust, or ``disentangled" representation~\cite{chen2016infogan} of the system under which models can generalize. This idea is active in computer vision, where learning based on invariance has become popular~\cite{krueger2020out}. Using similar ideas, we present a novel architecture for learning invariant representations for dynamics models.
% All these methods could be used to generalize our nominal model; in this paper we present a novel architecture for learning an invariant representation.

\section{METHOD}
\label{sec:methods}
Our approach is composed of two components: offline representation learning and online control. First, we present how we learn a representation that allows for generalization by exploiting inherent invariances inferred from the nominal data, shown in Fig.~\ref{fig:invariant tsf architecture}. 
Second, we present \controllername{} (\textbf{\controllerabv{}}), a two-level hierarchical MPC method shown in Fig.~\ref{fig:control architecture}. The high-level controller explicitly reasons about non-nominal dynamics and traps, deciding when to exploit the dynamics and when to recover to familiar ones by outputting the model and cost function the low-level controller uses to compute control signals.

\subsection{Offline: Invariant Representation for Dynamics}\label{sec:dynamics learning}
\begin{figure}
    \centering
    \def\svgwidth{0.49\textwidth}
    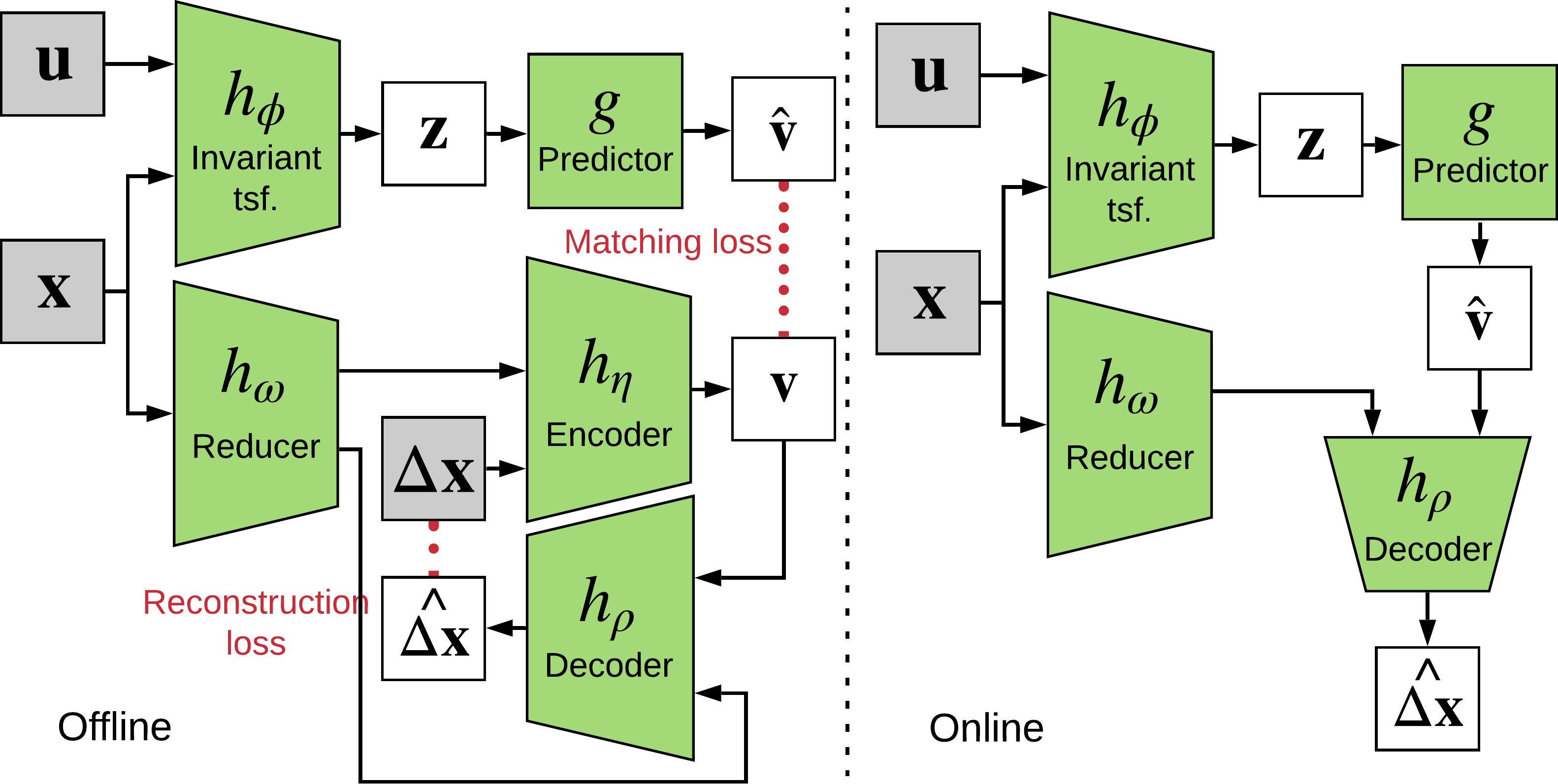
  \caption{Architecture for learning (left) and using (right) an invariant representation $\learnednom$. Grey = given data, green = parameterized transforms, white = computed values, and red dotted lines = losses.}
  \label{fig:invariant tsf architecture}
\end{figure}
In this section, our objective is to learn $\learnednom$ while exploiting potential underlying invariances in $\xall \times \usall$ to achieve better generalization to unseen data. More formally, our representation consists of an invariant transform $\encoder$ and a predictive module $\approxlatentdynamics$, shown in Fig.~\ref{fig:invariant tsf architecture}. $\encoder$ maps $\xall \times \usall$ to a latent space ($\z \in \reals^\nz$) that $\approxlatentdynamics$ maps to \vname\ ($\vs \in \reals^\nv$) that is then mapped back to $\xall$ using $\decoder$. We parameterize the transforms with neural networks and build in two mechanisms to promote meaningful latent spaces:

First, we impose $\nz < \nx + \nus$ to create an information bottleneck which encourages $\z$ to ignore information not relevant for predicting dynamics.
%, thus discovering invariances--variations in $\xall \times \usall$ that can be safely ignored. 
Typically, $\nz$ can be iteratively decreased until predictive performance on $\nomtraj$ drops significantly compared to a model in the original space.
% This encourages the \zname{} to throw out information not relevant for predicting dynamics and to discover invariances---variations in the original state-action space that can be safely ignored.
Further, we limit the size of $\approxlatentdynamics$ to be smaller than that of $\encoder$ such that the dynamics take on a simple form. 

Second, to reduce compounding errors when $\x,\us$ is OOD, we partially decouple $\z$ and $\vs$ by learning $\vs$ in an autoencoder fashion from $\x,\dx$ with encoder $\inversedecoder$ and decoder $\decoder$. 
% we require that $\vs$ can reconstruct $\dx$ with partial information from $\x$. 
% Passing in $\x$ at reconstruction allows the dynamics pathway to ignore some information in $\x$. 
Our innovation is to match the encoded $\vs$ with the $\hat{\vs}$ output from the dynamics predictor.
% Much like the autoencoder architecture, we require the encoded $\vs$ to match the $\hat{\vs}$ output from dynamics. 
To further improve generalization, we restrict information passed from $\x$ to $\vs$ with a dimension reducing transform $\extractor: \xall \rightarrow \reals^\nxtract$. 
%This mechanism has the effect of reducing compounding errors when $\x,\us$ is OOD. 
These two mechanisms yield the following expressions:
\begin{align*}
    \dx &\approx \hat{\Delta \x} = \decoder(\vs, \extractor(\x)) &
    \vs &= \inversedecoder(\dx, \extractor(\x)) \approx \hat{\vs} = \approxlatentdynamics(\z)
\end{align*}
and their associated batch reconstruction and matching loss:
\begin{align*}
    \lreconstruct &= \frac{\Expect{\norm{\dx - \decoder(\vs, \extractor(\x))}_2}} {\Expect{\norm{\dx}_2}} &
    \lmatch &= \frac{\Expect{\norm{\vs - \hat{\vs}}_2}} {\Expect{\norm{\vs}_2}}\\
    \lbase(\nomtraj_i) &= \weightreconstruct \lreconstruct(\nomtraj_i) +
    \weightmatch \lmatch(\nomtraj_i)
\end{align*}
These losses are ratios relative to the norm of the quantity we are trying to match to avoid decreasing loss by scaling the representation. In addition to these two losses,
we apply Variance Risk Extrapolation (V-REx \cite{krueger2020out}) to explicitly penalize the variance in loss across the $\ntraj$ trajectories:
\begin{equation}\label{eq:invariant loss}
    \loss(\nomtraj) = \rexparam \var{\{ \lbase(\nomtraj_1), ..., \lbase(\nomtraj_\ntraj)\}} + \sum_{i=1}^\ntraj \lbase(\nomtraj_i)
\end{equation}
We train on Eq.~(\ref{eq:invariant loss}) using gradient descent.
%with an annealing strategy for $\rexparam$ suggested by \cite{krueger2020out}.
% For minibatches, we adjust Eq.~(\ref{eq:invariant loss}) to be over only the trajectories that are in the batch.

After learning the transforms, we replace $\approxlatentdynamics$ with a higher capacity model and fine-tune it on the nominal data with just $\lmatch$.
% Learning the invariant representation this way avoids compounding errors from OOD inputs while allowing our model to be robust to variations unnecessary for dynamics. 
For further details, please see App. C. 
Since we have no access to $\dx$ online, we pass $\hat{\vs}$ to $\decoder$ instead of $\vs$:
\begin{equation}
     \learnednom(\x, \us) = \decoder(\approxlatentdynamics(\encoder(\x,\us)), \extractor(\x)) 
\end{equation}

\subsection{Online: Trap-Aware MPC}

Online, we require a controller that has two important properties. First, it should incorporate strategies to escape from and avoid detected traps. Second, it should iteratively improve its dynamics representation, in particular when encountering previously unseen modes. To address these challenges, our approach uses a two-level hierarchical controller where the high-level controller is described in Alg.~\ref{alg:high level controller}.
% \todo{Brad: This algorithm is hard to follow. One cause is the complicated branching logic. Can you rewrite this to make it easier to follow? What I want to see is: if s is NOM: Do nominal MPC. if s is NONNOM do some model updating. if s is recovery do the recovery stuff. Let me know if you want to discuss} explicitly reasons about non-nominal dynamics and traps, outputting the dynamics model and cost function that the low level controller uses to compute control. This structure allows a variety of low-level MPC designs that can be specialized to the task or dynamics if domain knowledge is available. 

\begin{figure}[t]
    \centering
    \def\svgwidth{0.49\textwidth}
    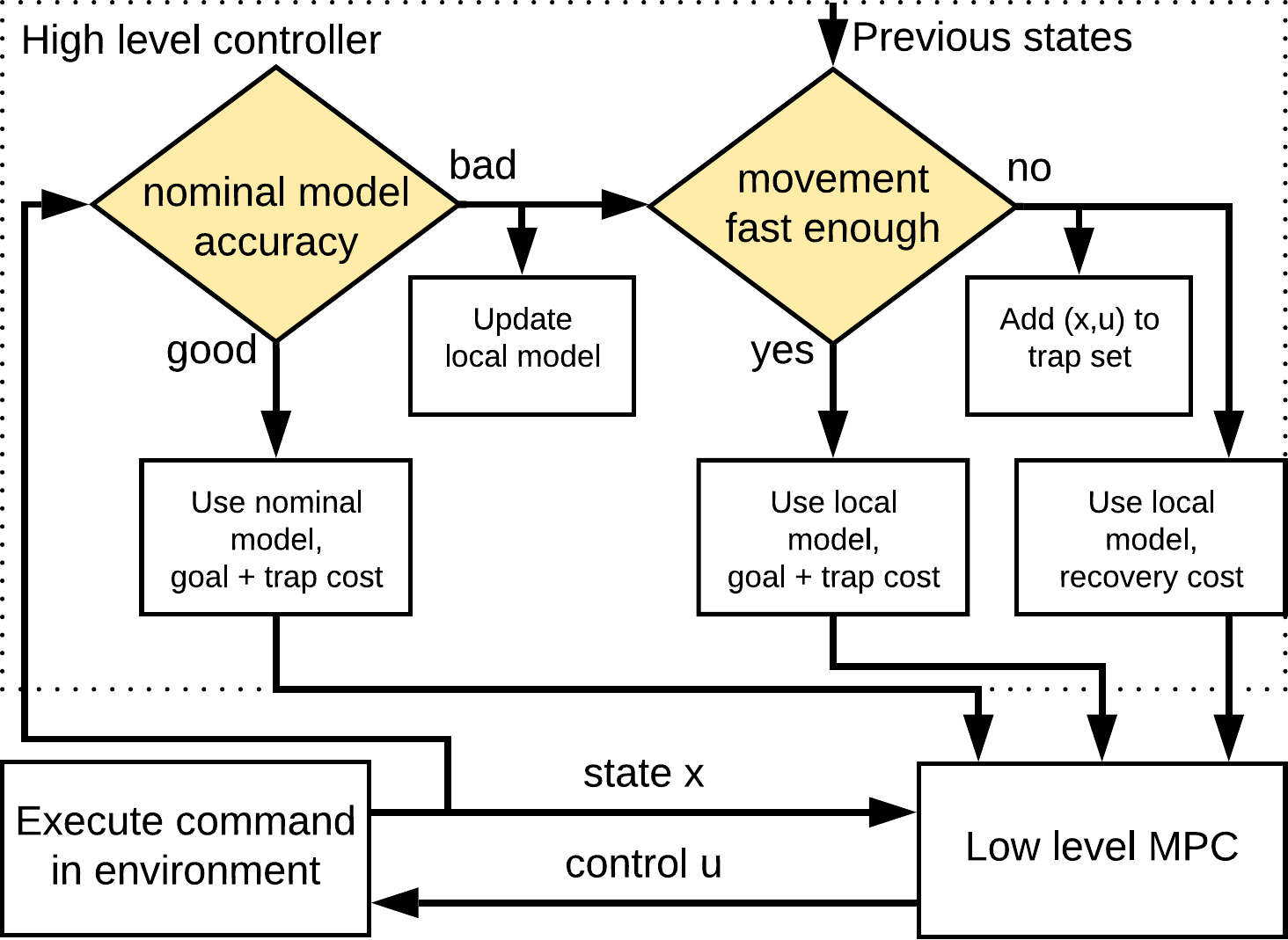
  \caption{High-level control architecture.}
  \label{fig:control architecture}
\end{figure}
% \begin{wrapfigure}[19]{R}{0.25\textwidth}
%   \includesvg[width=0.24\textwidth]{high level control arch.svg}
%   \caption{High-level control architecture.}
%   \label{fig:control architecture}
% \end{wrapfigure}

\controllerabv{} operates in either an exploit or recovery mode, depending on dynamics encountered. When in nominal dynamics, the algorithm exploits its confidence in predictions. When encountering non-nominal dynamics, it attempts to exploit a local approximation built online until it detects entering a trap, at which time recovery is triggered. This approach attempts to balance between a potentially overly-conservative treatment of all non-nominal dynamics as traps and an overly-optimistic approach of exploiting all non-nominal dynamics assuming goal progress is always possible.

The first step in striking this balance is identifying non-nominal dynamics. Here, we evaluate the nominal model prediction error against observed states (``nominal model accuracy" block in Fig.~\ref{fig:control architecture} and lines~\ref{line:nominal check 1} and~\ref{line:nominal check 2} from Alg.~\ref{alg:high level controller}):
\begin{equation}
    \norm{(\dx - \approxnominaldynamics(\x, \us))/E}_2 \ > \ \nominalthreshold
    \label{eq:nonnominal}
\end{equation} where $\nominalthreshold$ is a designed tolerance threshold and $E$ is the expected model error per dimension computed on the training data. %how much error we tolerate in nominal dynamics and $E$ is the expected error magnitude of each dimension. 
To handle jitter, we consider transitions from $\nnominal$ consecutive time steps.

When in non-nominal dynamics, the controller needs to differentiate between dynamics it can navigate to reach the goal vs. traps and adapt its dynamics models accordingly. 
Similar to~\cite{lee2003artificial}, we detect this as when we are unable to make progress towards the goal by considering the latest window of states. To be robust to cost function shape, we consider the state distance instead of cost.
Specifically, we monitor the maximum one-step state distance $\xdist(\x_t, \x_{t-1})$ in nominal dynamics, $\xvelnom$, and compare it against the average state distance to recent states: $\xdist(\x_t, \x_{a}) / (t - a) < \trapthreshold \xvelnom$
% \todo{Brad: Are you sure this is correct? suppose the robot encounters a trap late into execution so t is large but a is small. Then $t-a$ is large for a long long time.}
(depicted in ``movement fast enough" block of Fig.~\ref{fig:control architecture}). Here, $t$ is the time from task start. We consider $a=t_0,...,t-\ndynamics$, where $t_0$ is the start of non-nominal dynamics or the end of last recovery, whichever is more recent. We ensure state distances are measured over windows of at least size $\ndynamics$ to handle jitter. $\trapthreshold$ is how much slower the controller tolerates moving in non-nominal dynamics. For more details see Alg.~\ref{alg:enter trap}.

\begin{algorithm}[tb]
\DontPrintSemicolon
\SetKwInOut{Input}{Input}
\SetKwInOut{Given}{Given}
\SetKwInOut{Output}{Output}
\SetKwInOut{Hyperparameters}{Hyperparameters}

  \SetKwFunction{FE}{EnteringTrap}
  \SetKwFunction{NN}{NonNominal}
  \SetKwFunction{ATS}{AddToTrapSet}
  \SetKwFunction{FND}{LeftNonnominalDynamics}
  \SetKwFunction{FLR}{Recovered}
  \SetKwFunction{FER}{EndRecovery}
  \SetKwProg{Fn}{Function}{:}{}
  
\Given{$\costfunc(\x,\us)$ cost, 
$\x_0$,
$\texttt{MPC}$,
parameters from Tab.~\ref{tab:high level controller parameters}
}
$\ctrlmode \leftarrow $ \nominalmode,\ \
$t \leftarrow 0$,\ \
$\xvelnom \leftarrow 0$,\ \
$\x \leftarrow \x_0$,\ \
$\trapours \leftarrow \{\}$,\ \
$\spread \leftarrow 1$,\ \
$\mabweight \leftarrow \textbf{0}$\;
% $\spreadmin \leftarrow \maxcost \clearance^2$\label{line:param init}\;
MAB arms $\leftarrow \nmabarms$ random convex combs.\;
\While{$\costfunc(\x, 0) > $ acceptable threshold}{
    \uIf{\ctrlmode{} is \nominalmode} {
        \uIf{not nominal from Eq.~\ref{eq:nonnominal}\label{line:nominal check 1}}{ 
            $\ctrlmode \leftarrow$ \nonnominalmode\;
            initialize GP $\approxerrordynamics$ with $(\x_{t-1}, \us_{t-1}, \dx_{t-1})$\;
        } 
        \Else{
            $\spread \leftarrow \spread \cdot{} \trapanneal$ \label{line:anneal trap cost} \tcp{anneal}
            % $\spread \leftarrow \max(\spreadmin, \spread \cdot{} \trapanneal)$ \label{line:anneal trap cost} \tcp{anneal}
            $\xvelnom \leftarrow \max(\xvelnom, \xdist(\x_{t-\ndynamics}, \x)/\ndynamics)$ 
        }
    }
    \Else{
        fit $\approxerrordynamics$ to include $(\x_{t-1}, \us_{t-1}, \dx_{t-1})$\;
        $n \leftarrow$ was nominal last $\nnominal$ steps \label{line:nominal check 2} \tcp{Eq.~\ref{eq:nonnominal}}
        \uIf{\FE{$\xvelnom$}} {
            $\ctrlmode \leftarrow$ \recovermode\;
            expand $\trapours$ according to Eq.~\ref{eq:add to trapset}
            % \ATS{$\trapours$}\;
        }
        \uIf{\ctrlmode{} is \recovermode{}} {
            \uIf{n \textbf{or} \FLR{$\xvelnom$}} {
                $\ctrlmode \leftarrow$ \nonnominalmode\;
                $\spread \leftarrow$ normalize $\spread$ so $|\costtrap| \sim |\costfunc|$\;
            }
            \uElseIf{$\nmab$ steps since last arm pull}{
                reward last arm pulled with $\xdist(\x_t, \x_{t-\nmab}) / \nmab\xvelnom$\;
                $\mabweight \leftarrow $ Thompson sample an arm\;
            }
        }
        \uIf{n}{
            $\ctrlmode \leftarrow$ \nominalmode\;
        }
    }
    \texttt{MPC}.model $\leftarrow \approxnominaldynamics$ \textbf{if} \ctrlmode{} is \nominalmode{} \textbf{else} $\approxnominaldynamics + \approxerrordynamics$\;
    \texttt{MPC}.cost $\leftarrow \recoveryweight \cdot (\mabweight_1\costrecovery(\xnom) + \mabweight_2\costrecovery(\xfastest))$ \textbf{if} $\ctrlmode{}$ is $\recovermode{}$ \textbf{else} $\costfunc + \spread \costtrap$ \tcp{Eq.~\ref{eq:recovery} and \ref{eq:trap}}\label{line:recovery cost}
    
    $\us \leftarrow \texttt{MPC}(\x)$, $t \leftarrow t + 1$\;
    $\x \leftarrow $ apply $\us$ and observe from env
}
\caption{\controllerabv{} high-level control loop}
\label{alg:high level controller}
\end{algorithm}

Our model adaptation strategy, for both non-nominal dynamics and traps, is to mix the nominal model with an online fitted local model. Rather than the linear models considered in prior work~\cite{fu2016one}, we add an estimate of the error dynamics $\approxerrordynamics$ represented as a Gaussian Process (GP) to the output of the nominal model.
% fit a Gaussian Process (GP) $\approxerrordynamics$ with $\approxlatentdynamics$ as the mean function to estimate error dynamics.
Using a GP provides a sample-efficient model that captures non-linear dynamics. To mitigate over-generalizing local non-nominal dynamics to where nominal dynamics holds, we fit it to only the last $\nlocal$ points since entering non-nominal dynamics. We also avoid over-generalizing the invariance that holds in nominal dynamics by constructing the GP model in the original state-control space. Our total dynamics is then 
\begin{align}
 \approxnewdynamics(\x, \us) &= \approxnominaldynamics(\x, \us)+ \approxerrordynamics(\x, \us)\\
 \newdynamics(\x,\us) &\approx \Expect[\approxnewdynamics(\x,\us)] \label{eq:new approx}
\end{align}
When exploiting dynamics to navigate towards the goal, we regularize the goal-directed cost $\costfunc$ with a trap set cost $\costtrap$ to avoid previously seen traps (line~\ref{line:recovery cost} from Alg.~\ref{alg:high level controller}). This trap set $\trapours$ is expanded whenever we detect entering a trap. We add to it the transition with the lowest ratio of actual to expected movement (from one-step prediction, $\hat{\x}$) since the end of last recovery: 
%$\tau \leftarrow \min_a \xdist(\x_a, \x_{a+1}) / \xdist(\x_a, \hat{\x}_{a+1})$, 
%$\trapours \leftarrow \trapours \cup \{(\x_{\tau}, \us_{\tau})\}$.
% \begin{equation}
%     \tau = \min_a \frac{\xdist(\x_a, \x_{a+1})}{\xdist(\x_a, \hat{\x}_{a+1})},
% \trapours \leftarrow \trapours \cup \{(\x_{\tau}, \us_{\tau})\}
% \label{eq:add to trapset}
% \end{equation}
\begin{equation}
    b = \argmin_a \frac{\xdist(\x_a, \x_{a+1})}{\xdist(\x_a, \hat{\x}_{a+1})},
\trapours \leftarrow \trapours \cup \{(\x_{b}, \us_{b})\}
\label{eq:add to trapset}
\end{equation}
To handle traps close to the goal, we only penalize revisiting trap states if similar actions are to be taken. With the control similarity function $\usim : \usall \times \usall \rightarrow [0,1]$ we formulate the cost, similar to the virtual obstacles of~\cite{lee2003artificial}:
\begin{equation}
    \costtrap(\x, \us) = \sum_{\x', \us' \in \trapours} {\frac{\usim(\us, \us')} {\xdist(\x, \x')^2}}\label{eq:trap}
\end{equation}
The costs are combined as
$\costfunc + \spread \costtrap$ (line~\ref{line:recovery cost} from Alg.~\ref{alg:high level controller}), where $\spread \in (0,\infty)$
is annealed by $\trapanneal \in (0, 1)$ each step in nominal dynamics (line~\ref{line:anneal trap cost} from Alg.~\ref{alg:high level controller}).
%, down to a minimum of $\spreadmin = \maxcost \clearance^2$.
% Clearance $\clearance \in (0,\infty)$ relates to the minimum distance (in $\xdist$) from traps we wish to be. 
Annealing the cost avoids assigning a fixed radius to the traps, which if small results in inefficiency as we encounter many adjacent traps, and if large results in removing many paths to the goal set.
% Assuming the control similarity function has positive measure support, the minimum multiplier $\spreadmin$ ensures that the \textbf{removed set}
% $\removedset = \{\x, \us \ |\  \spread \costtrap(\x, \us) \ge \maxcost\}$ will have positive measure. A controller optimizing $\costfunc + \spread \costtrap$ will not plan to enter $\removedset$.
% $\costtrap(\x,\us') \ge \maxcost \ \forall \xdist(\x, \x') \le \clearance, \ \x',\us' \in \trapours$

We switch from exploit to recovery mode when detecting a trap, but it is not obvious what the recovery policy should be. Driven by the online setting and our objective of data-efficiency: %, we inject structure into the policy.
First, we restrict the recovery policy to be one induced by running the low-level MPC on some cost function other than one used in exploit mode.
Second, we propose hypothesis cost functions and consider only convex combinations of them. 
Without domain knowledge, one hypothesis is to return to one of the last visited nominal states. However, the dynamics may not always allow this. Another hypothesis is to return to a state that allowed for the most one-step movement. Both of these are implemented in terms of the following cost, where $S$ is a state set and we pass in either $\xnom$, the set of last visited nominal states, or $\xfastest$, the set of $\nfastest$ states that allowed for the most single step movement since entering non-nominal dynamics:
\begin{equation}
    \costrecovery(\x, \us, S) = \min_{\x' \in S} {d(\x, \x')^2}\label{eq:recovery}
\end{equation}
Third, we formulate learning the recovery policy online as a non-stationary multi-arm bandit (MAB) problem. We initialize $\nmabarms$ bandit arms, each a random convex combination of our hypothesis cost functions. Every $\nmab$ steps in recovery mode, we pull an arm to select and execute a recovery policy. After executing $\nmab$ control steps, we update that arm's estimated value with a movement reward: $\xdist(\x_t, \x_{t-\nmab}) / \nmab\xvelnom$. When in a trap, we assume any movement is good, even away from the goal. The normalization makes tuning easier across environments.
To accelerate learning, we exploit the correlation between arms, calculated as the cosine similarity between the cost weights. 
Our formulation fits the problem from \cite{mcconachie2020bandit} and we implement their framework for non-stationary correlated multi-arm bandits. 

% Ignoring data efficiency, a fully parameterized policy $\us = \pi(\x)$ could be learned in a reinforcement learning manner online with some form of long-horizon movement as reward, but since we are doing online control we inject more structure to the policy and thus require far less data. 

Finally, we return to exploit mode after a fixed number of steps $\nrecoverymax$, if we returned to nominal dynamics, or if we stopped moving after leaving the initial trap state. For details see Alg.~\ref{alg:leave recovery}.

\section{EXPERIMENTS}
\label{sec:results}

\newcommand{\sidelength}{a}
\newcommand{\yaw}{\theta}
\newcommand{\reaction}{r}
\newcommand{\pushmag}{\delta}
\newcommand{\pushdir}{\alpha}
\newcommand{\pushalong}{p}
\newcommand{\rot}[1]{R(#1)}
\newcommand{\cossim}{\text{cossim}}

In this section, we first evaluate our dynamics representation learning approach, in particular how well it generalizes out-of-distribution. Second, we compare \controllerabv{} against baselines on tasks with traps in two environments.

% Similar to our method section, we split results into offline and online components. First, we evaluate the offline learning of a dynamics representation on whether it generalizes out-of-distribution. Second, we compare \controllerabv{} against baselines on tasks with traps in simulated environments.

\subsection{Experiment Environments}
\begin{figure}
  \centering
  \includegraphics[width=0.49\textwidth]{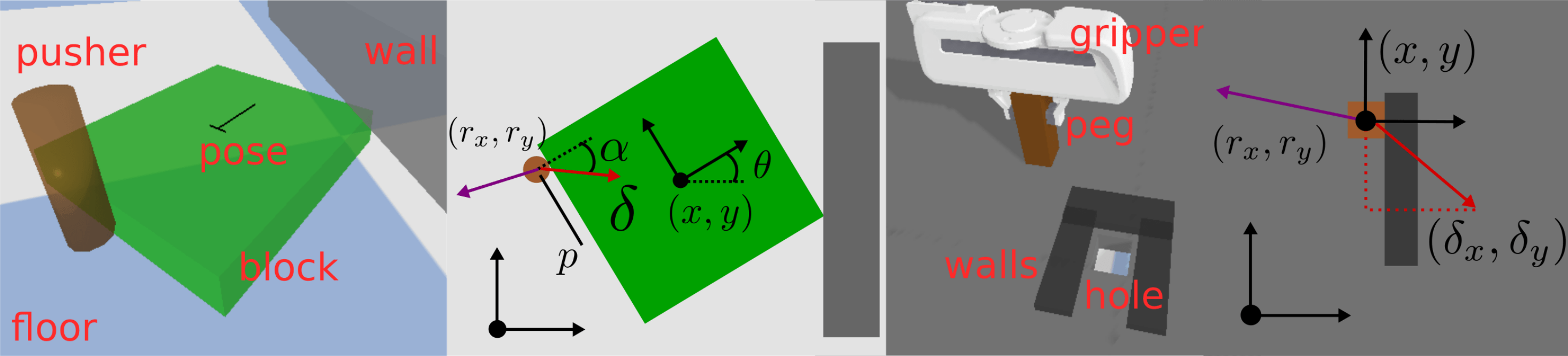}
  \caption{Annotated simulation environments of (left) planar pushing, and (right) peg-in-hole.}
  \label{fig:env}
\end{figure}
Our two tasks are quasi-static planar pushing and peg-in-hole. Both tasks are evaluated in simulation using PyBullet ~\cite{coumans2016pybullet} and the latter is additionally evaluated empirically using a 7DoF KUKA LBR iiwa arm depicted in Fig.~\ref{fig:real peg tasks}. 
Our simulation time step is 1/240s, however each control step waits until reaction forces are stable. The action size for each control step is described in App B.
In planar pushing, the goal is to push a block to a known desired position. In peg-in-hole, the goal is to place the peg into a hole with approximately known location. In both environments, the robot has access to its own pose and senses the reaction force at the end-effector. \textbf{ Thus the robot cannot perceive the obstacle geometry visually, it only perceives contact through reaction force.} \textbf{During offline learning of nominal dynamics, there are no obstacles or traps}. During online task completion, obstacles are introduced in the environment, inducing unforeseen traps. See Fig.~\ref{fig:env} for a depiction of the environments and Fig.~\ref{fig:task init} for typical traps from tasks in these environments, and App. B for environment details.

In planar pushing, the robot controls a cylindrical pusher restricted to push a square block from a fixed side. Fig.~\ref{fig:task init} shows traps introduced by walls. Frictional contact with a wall limits sliding along it and causes most actions to rotate the block into the wall. 
State is $\x = (x,y,\yaw,\reaction_x,\reaction_y)$ where $(x,y,\yaw)$ is the block pose, and $(\reaction_x, \reaction_y)$ is the reaction force the pusher feels, both in world frame.
Control is $\us = (\pushalong, \pushmag, \pushdir)$, where $\pushalong$ is where along the side to push, $\pushmag$ is the push distance, and $\pushdir$ is the push direction relative to side normal.
The state distance is the 2-norm of the pose, with yaw normalized by the block's radius of gyration. The control similarity is $\usim(\us_1, \us_2) = \max(0, \cossim((\pushalong_1,\pushdir_1), (\pushalong_2,\pushdir_2)))$ where $\cossim$ is cosine similarity.

In peg-in-hole, we control a gripper holding a peg (square in simulation and circular on the real robot) that is constrained to slide along a surface. Traps in this environment geometrically block the shortest path to the hole. The state is $\x = (x,y,z,\reaction_x,\reaction_y)$ and control is $\us = (\pushmag_x, \pushmag_y)$, the distance to move in $x$ and $y$. We execute these on the real robot using a Cartesian impedance controller.
The state distance is the 2-norm of the position and the control similarity is $\usim(\us_1, \us_2) = \max(0, \cossim(\us_1, \us_2))$. The goal-directed cost for both environments is in the form $\costfunc(\x,\us) = \x^TQ\x + \us^T R\us$. 
The MPC assigns a terminal multiplier of 50 at the end of the horizon on the state cost. 
See Tab.~\ref{tab:cost parameters} for the cost parameters of each environment.

% The challenge for planar pushing is that the traps require long action sequences to escape from. While traps are easier to escape in peg-in-hole, we explore sophisticated trap shapes and when they are near the goal.

$\nomtraj$ for simulated environments consists of $\ntraj=200$ trajectories with $\nsteps=50$ transitions (all collision-free). For the real robot, we use $\ntraj=20$, $\nsteps=30$. We generate them by uniform randomly sampling starts from $[-1,1] \times [-1,1]$ ($\yaw$ for planar pushing is also uniformly sampled; for the real robot we start each randomly inside the robot workspace) and applying actions uniformly sampled from $\usall$.

\subsection{Offline: Learning Invariant Representations}
\begin{figure}[t]
\includegraphics[width=\linewidth]{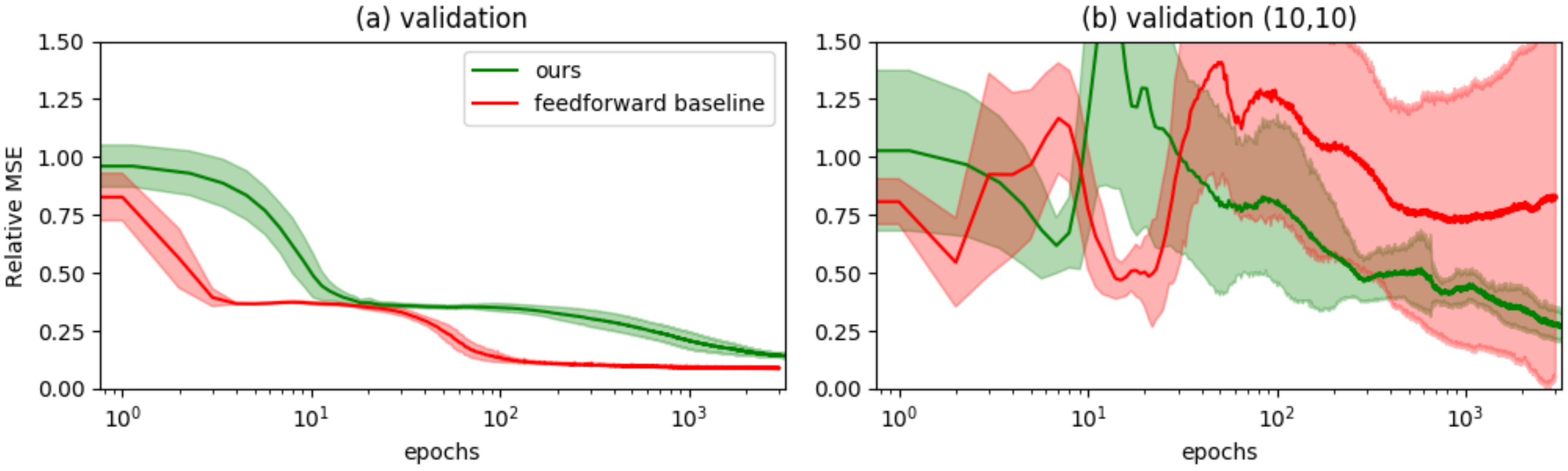}
%   \caption{Losses on different datasets to test out the learned representation's generalization ability. Apart from the validation set (left), the other datasets are all out-of-distribution.}
   \caption{Learning curves on validation and OOD data sets for planar pushing representation. Mean across 10 runs is solid while 1 std. is shaded.}
  \label{fig:learning}
%   \vspace{-2em}
\end{figure}

In this section we evaluate if our representation can learn useful invariances from offline training on $\nomtraj$. We expect nominal dynamics in freespace in our environments to be invariant to translation.
Since $\nomtraj$ has positions around $[-1,1] \times [-1,1]$, we evaluate translational invariance translating the validation set by $(10,10)$. We evaluate relative MSE $\norm{\dxe - \dx}_2/\Expect{\norm{\dx}_2}$ (we do not directly optimize this) against a fully connected baseline of comparable size mapping $\x,\us$ to $\hat{\Delta \x}$ learned on relative MSE.
As Fig.~\ref{fig:learning}b shows, our performance on the translated set is better than the baseline, and trends toward the original set's performance. Note that we expect our method to have higher validation MSE since V-REx sacrifices in-distribution loss for lower variance across trajectories.
% we achieve good performance on the translated validation set even with ablations learned without REx. 
% This could be due to our low dimensional state space whereas REx was developed for high dimensional image representations. 
We use $\nz = 5, \nv = 5, \nxtract = 2$, and implement the transforms with fully connected networks. For network sizes and learning details see App. C.

% Fig.~\ref{fig:learning}c shows performance on a test set with OOD reaction forces \jznote{(about 50\% of data is in contact with an obstacle)}, which we do not expect invariance over. Since the nominal data has no obstacles, we expect dynamics to do poorly.\todo{since we're comparing MSE now, can remove test set results} % do reconstruction\todo{"reconstruction" is ambiguous here, clarify}.

\subsection{Online: Tasks in Novel Environments}
\begin{figure}[t]
\includegraphics[width=\linewidth]{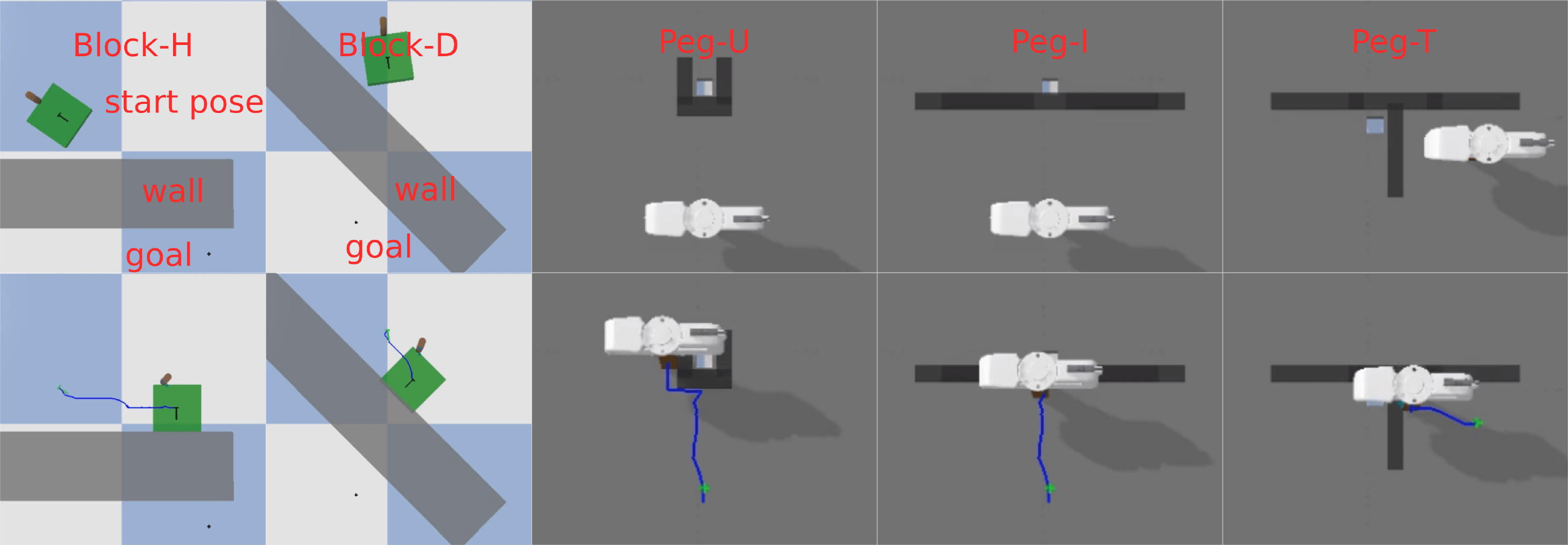}
   \caption{(top) Initial condition and (bottom) typical traps for planar pushing and peg-in-hole tasks. Our method has no visual sensing and is pre-trained only on environments with no obstacles.}
  \label{fig:task init}
  \vspace{-6pt}
\end{figure}
We evaluate \controllerabv{} against baselines and ablations on the tasks shown in Fig.~\ref{fig:real peg tasks} and Fig.~\ref{fig:task init}. 
%We emphasize that 1) we assume no visual sensing and thus can only infer obstacles through contact; and 2) our offline training does not include any obstacles or traps of any kind. 
For \controllerabv{}'s low-level MPC, we use a modified model predictive path integral controller (MPPI)~\cite{williams2017information} where we take the expected cost across $\nrollouts=10$ rollouts for each control trajectory sample to account for stochastic dynamics. 
See Alg.~\ref{alg:mppi} for our modifications to MPPI. \controllerabv{} takes less than 1s to compute each control step for all tasks. We run for 500 steps (300 for Real Peg-T).

\textbf{Baselines:}
We compare against five baselines. The first is the APF-VO method from~\cite{lee2003artificial}, which uses the gradient on an APF to select actions. The potential field is populated with repulsive balls in $\xall$ based on $\xdist$ as we encounter local minima. We estimate the gradient by sampling 5000 single-step actions and feeding through $\learnednom$. Second is an APF-LME method from~\cite{fedele2018obstacles} (APF-SP) using switched potentials between the attractive global potential, and a helicoid obstacle potential, suitable for 2D obstacles.
%We also try a variant of the APF-LME where we online estimate error dynamics using the GP. 
The APF methods use the $\learnednom$ learned with our proposed method (Section~\ref{sec:dynamics learning}) for next-state prediction.
Third is online adaptive MPC from~\cite{fu2016one} (``adaptive MPC++"), which does MPC on a linearized global model mixed with a locally-fitted linear model. 
%This represents the overly-optimistic approach of always exploiting non-nominal dynamics to head towards the goal.
iLQR (code provided by~\cite{fu2016one}'s author) performs poorly in freespace of the planar pushing environment. We instead use MPPI with a locally-fitted GP model (effectively an ablation of \controllerabv{} with control mode fixed to $\nonnominalmode$). 
Next is model-free reinforcement learning with Soft Actor-Critic (SAC)~\cite{haarnoja2018soft}. Here, a nominal policy is learned offline for 1000000 steps on the nominal environment, which is used to initialize the policy at test time. Online, the policy is retrained after every control on the dense environment reward. 
Lastly, our ``non-adaptive'' baseline runs MPPI on the nominal model.

We also evaluated ablations to demonstrate the value of \controllerabv{} components. ``\controllerabv{} rand. rec." takes uniform random actions until dynamics is nominal instead of using our recovery policy. ``\controllerabv{} original space" uses a dynamics model learned in the original $\xall \times \usall$ (only for Peg-T(T)). Lastly, ``\controllerabv{} $e=0$" does not estimate error dynamics. 

\begin{figure*}[t]
\includegraphics[width=\linewidth]{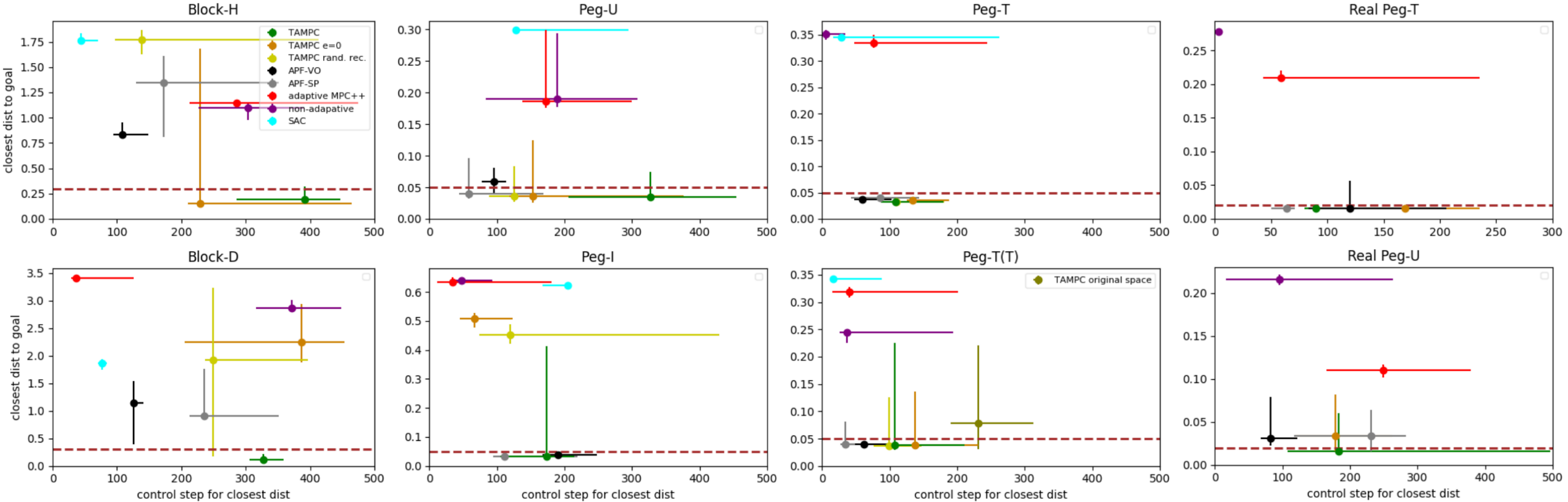}
   \caption{Minimum distance to goal after 500 steps (300 for Real Peg-T) accounting for walls (computed with Djikstra's algorithm). Median across 10 runs is plotted with error bars representing 20--80\textsuperscript{th} percentile. Task success is achieving lower distance than the dotted red line.}
    \vspace{-2em}
  \label{fig:task res}
\end{figure*}

\begin{table}[]
\caption{Success counts across all tasks, as defined by achieving distance below the thresholds in Fig.~\ref{fig:task res}.}\label{tab:res}
\tabcolsep=0.12cm
\begin{tabular}{|l|ll|llll|ll|}
\hline
\multirow{2}{*}{Method} & \multicolumn{8}{l|}{Success over 10 trials}        \\ \cline{2-9} 
                        & B-H & B-D & P-U & P-I & P-T & P-T(T) & RP-T & RP-U \\ \hline
\textbf{TAMPC}          & 8   & 9   & 7   & 7   & 9   & 6      & 10   & 6    \\
TAMPC e=0               & 6   & 1   & 5   & 0   & 9   & 7      & 9    & 3    \\
TAMPC rand. rec.            & 1   & 4   & 5   & 0   & 9   & 7      & -    & -    \\ \hline
APF-VO                  & 0   & 1   & 4   & 10  & 10  & 10     & 8    & 2    \\
APF-SP                  & 1   & 0   & 5   & 10  & 10  & 8      & 10   & 4    \\ \hline
adaptive MPC++          & 0   & 0   & 0   & 0   & 0   & 0      & 0    & 0    \\
non-adaptive            & 0   & 0   & 0   & 0   & 0   & 0      & 0    & 0    \\
SAC                     & 0   & 0   & 0   & 0   & 0   & 0      & -    & -    \\ \hline
\end{tabular}
\end{table}

% \section{DISCUSSION}
\subsection{Task Performance Analysis}\label{sec:task performance}
Fig.~\ref{fig:task res} and Tab.~\ref{tab:res} summarizes the results. For Fig.~\ref{fig:task res}, ideal controller performance would approach the lower left corners. We see that \controllerabv{} outperforms all baselines on the block pushing tasks and Peg-U, with slightly worse performance on the other peg tasks compared to APF baselines. APF baselines struggled with block pushing 
% due to the dynamics requiring multiple control steps to escape contact with the wall, and the inability of APF methods to plan ahead. Instead, they must iteratively fill up local minima, which is inefficient and accounts for their low success rates. Additionally, 
since turning when the goal is behind the block is also a trap for APF methods, because any action increases immediate cost and they do not plan ahead. On the real robot, joint limits were handled naturally as traps by \controllerabv{} and APF-VO.% the Cartesian impedance controller did not account for joint limits; instead, they were handled naturally as traps by \controllerabv{} and APF-VO.} 
% shows that TAMPC or its variants outperforms baselines on all tasks in median distance to goal after 500 control steps. 

Note that the APF baselines were tuned to each task, thus it is fair to compare against tuning \controllerabv{} to each task. However, we highlight that 
our method is empirically robust to parameter value choices, as we achieve high success even when using the same parameter values across tasks and environments, listed in Tab.~\ref{tab:high level controller parameters}. Peg-U and Peg-I were difficult tasks that benefited from independently tuning \textit{only three} important parameters, which we give intuition for: We control the exploration of non-nominal dynamics with $\ndynamics$. For cases like Peg-U where the goal is surrounded by non-nominal dynamics, we increase exploration by increasing $\ndynamics$ with the trade-off of staying longer in traps. 
Independently, we control the expected trap basin depth (steps required to escape traps) with the MPC horizon $\horizon$.
Intuitively, we increase $\horizon$ to match deeper basins, as in Peg-I, at the cost of more computation.
Lastly, $\trapanneal \in (0,1)$ controls the trap cost annealing rate. Too low a value prevents escape from difficult traps while values close to 1 leads to waiting longer in cost function local minima.
We used $\ndynamics=15, \horizon=15$ for Peg-U, and $\horizon=20, \trapanneal=0.95$ for Peg-I.

For APF-VO, Peg-U was difficult as the narrow top of the U could be blocked off if the square peg caught on a corner at the entrance. In these cases, \controllerabv{} was able to revisit close to the trap state by applying dissimilar actions to before. This was less an issue in Real Peg-U as we used a round peg, but a different complicating factor is that the walls are thinner (compared to simulation) relative to single-step action size. This meant that virtual obstacles were placed even closer to the goal.
APF-SP often oscillated in Peg-U due to traps on either side of the U while inside it.
% On the Real Peg-U, oscillations were escaped due to the accumulated deflection of the peg in the gripper from repeatedly pushing into walls.}
% See Appendix~\ref{ap:task tuning} for tuning details.

% On the real robot, the Cartesian impedance controller did not account for joint limits; instead, these limits were handled naturally as traps by \controllerabv{}. The only scenario in which the baselines achieved some level of success was the real Peg-U, where these approaches generated controls signals that always ended up sliding the peg along the same side of the U. This may be due to possibly imperfect construction leading one corner of the obstacle to be be closer to the goal, or that the joint configuration favoured moving in that direction. This movement bias may have contributed to the adaptive baseline's success. In contrast, the non-adaptive baseline did not exhibit any movement biases and stayed at the bottom of the U, while \controllerabv{} trajectories explored both sides.

% A common trend we identify from Fig.~\ref{fig:task res} is that the baselines tend to plateau after initially decreasing distance to goal. Through inspection, we found that this plateau did indeed correspond to being caught in traps. 
%with the low variance across runs (apart from Peg-U) suggesting they stay in the same trap.

The non-\controllerabv{} and non-APF baselines tend to cluster around the top left corner in Fig.~\ref{fig:task res}, indicating that they entered a trap quickly and never escaped. Indeed, we see that they all never succeed. For adaptive MPC, this may be due to a combination of insufficient knowledge of dynamics around traps, over-generalization of trap dynamics, and using too short of a MPC horizon. 
SAC likely stays inside of traps because no action immediately decreases cost, and action sequences that eventually decrease cost have high short-term cost and are thus unlikely to be sampled in 500 steps. 
% \controllerabv{} escapes traps because it uses the signal from state distance to explicitly reason about traps and switch to a recovery policy.
% This added structure is useful in tasks with traps because it decreases the degree of learning and planning the algorithm has to do. 
% \jznote{We can interpret this as injecting a prior that traps may exist and should be escaped from differently than going towards the goal.}

\subsection{Ablation Studies}
% \jzc{did} not use the tuned parameters\todo{D: it's unclear what is meant be "tuned parameters" since there is no independent "tuned" baseline. Instead I would say that the performance may be improved by further tuning these methods: i.e. if you haven't done the tests you can't say for sure if performance would be better or worse but you can conjecture. You didn't use the best tuning, but you did use reasonable parameters. for Peg-U and Peg-I, \jzc{explaining} their decreased performance on \jzc{them}.} 
Pushing task performance degraded on both \controllerabv{} rand. rec. and \controllerabv{} $e=0$, suggesting value from both the recovery policy and local error estimation. This is likely because trap escape requires long action sequences exploiting the local non-nominal dynamics to rotate the block against the wall. This is unlike the peg environment where the gripper can directly move away from the wall and where \controllerabv{} rand. rec. performs well. Note that ablations used the parameter values in Tab.~\ref{tab:high level controller parameters} for Peg-I and Peg-U, instead of the tuned parameters from Section~\ref{sec:task performance}. This may explain their decreased performance on them. The Peg-T(T) task (copy of Peg-T translated 10 units in $x,y$) highlights our learned dynamics representation. Using our representation, we maintain closer performance to Peg-T than \controllerabv{} original space (3 successes). This is because annealing the trap set cost requires being in recognized nominal dynamics, without which it is easy to get stuck in local minima.
%Note that for tasks Peg-U, Peg-I, and Peg-T(T), at least 20\% of runs do not reach the goal. 

\section{CONCLUSION}\label{sec:conclusion}
We presented \controllerabv{}, a controller that escapes traps in novel environments, and showed that it performs well on a variety of tasks with traps in simulation and on a real robot. Specifically, it is capable of handling cases where traps are close to the goal, and when the system dynamics require many control steps to escape traps. In contrast, we showed that trap-handling baselines struggle in these scenarios. Additionally, we presented and validated a novel approach to learning an invariant representation. Through the ablation studies, we demonstrated the individual value of \controllerabv{} components: learning an invariant representation of dynamics to generalize to out-of-distribution data, estimating error dynamics online with a local model, and executing trap recovery with a multi-arm-bandit based policy.
Finally, the failure of adaptive control and reinforcement learning baselines on our tasks suggests that it is beneficial to explicitly consider traps. Future work will explore higher-dimensional problems where the state space could be computationally challenging for the local GP model. 

% \subsection{Limitations and Future Work}
% The main failure mode of \controllerabv{} is oscillation between previously visited traps. Future work could break their symmetry by weighing traps on time of visit. A limitation of our method is that we need to be given a state space with a distance function. Future work could attempt to learn this from nominal data, for example by assuming linearity of state distance to control effort. Our experiments focus on rigid-body interactions that can be succinctly represented with low-dimensional state-spaces. Future work could apply \controllerabv{} to non-rigid manipulation problems, where the higher dimensional state space could be computationally challenging for the local GP model. 

% Finally, future work could improve the detection of non-nominal dynamics. For certain classes of problems, there could be more justified measures than the 2-norm on model prediction error.

\appendices
\renewcommand{\thesectiondis}[2]{\Alph{section}:}

\section{Parameters and Algorithm Details}
\begin{table}[tb]
\caption{MPPI parameters for different environments.}
\label{tab:mppi parameters}
\centering

\begin{tabular}{|l|c|c|c|}
\hline
Parameter & block & peg & real peg \\ \hline
$\samples$ samples & 500 & 500 & 500\\
$\horizon$ horizon & 40 & 10 & 15\\
$\nrollouts$ rollouts & 10 & 10 & 10\\
$\lambda$ & 0.01 & 0.01 & 0.01\\
$\us'$ & $[0, 0.5, 0]$ & $[0, 0]$ & $[0, 0]$\\
$\mu$ & $[0, 0.1, 0]$ & $[0, 0]$ & $[0, 0]$\\
$\Sigma$ & $\diag{[0.2, 0.4, 0.7]}$ & $\diag{[0.2, 0.2]}$ & $\diag{[0.2, 0.2]}$\\ \hline 
\end{tabular}
\end{table}
Making U-turns in planar pushing requires $H \ge 25$. We shorten the horizon to 5 and remove the terminal state cost of 50 when in recovery mode to encourage immediate progress.

\begin{table}[tb]
\caption{\controllerabv{} parameters across environments.}
\vspace{-8pt}
\label{tab:high level controller parameters}
\begin{center}
\begin{tabular}{|l|c|c|c|}
\hline
Parameter & block & peg & real peg \\ \hline
$\trapanneal$ trap cost annealing rate & 0.97 & 0.9 & 0.8\\
$\recoveryweight$ recovery cost weight & 2000 & 1 & 1\\
$\nominalthreshold$ nominal error tolerance & 8.77 & 12.3 & 15\\
$\trapthreshold$ trap tolerance & 0.6 & 0.6 & 0.6\\
$\ndynamics$ min dynamics window & 5 & 5 & 2\\
$\nnominal$ nominal window & 3 & 3 & 3\\
$\nmab$ steps for bandit arm pulls & 3 & 3 & 3\\
$\nmabarms$ number of bandit arms & 100 & 100 & 100\\
$\nrecoverymax$ max steps for recovery & 20 & 20 & 20\\
$\nlocal$ local model window & 50 & 50 & 50\\
$\convergedthreshold$ converged threshold & 0.05$\trapthreshold$ & 0.05$\trapthreshold$ & 0.05$\trapthreshold$\\
$\movedthreshold$ move threshold & 1 & 1 & 1\\
% $\clearance$ clearance & 1e-8 & 1e-8 & 1e-8\\
\hline
\end{tabular}
\end{center}
\end{table}
% Block pushing has the default value of $\trapanneal$ because the dynamics prevent us from returning to a previously visited trap most of the time. Peg-in-hole benefits from a higher annealing rate (lower value of $\trapanneal$) so the robot spends less time in local minima formed by the conflicting attractive potential field from the goal and the repulsive potential field from the trap set cost. 

$\recoveryweight \in (0,\infty)$ depends on how accurately we need to model trap dynamics to escape them. Increasing $\recoveryweight$ leads to selecting only the best sampled action while a lower value leads to more exploration by taking sub-optimal actions.
% mixing from less optimal actions, \jznote{allowing} more exploration.

Nominal error tolerance $\nominalthreshold$ depends on the variance of the model prediction error in nominal dynamics. A higher variance requires a higher $\nominalthreshold$. We use a higher value for peg-in-hole because of simulation quirks in measuring reaction force from the two fingers gripping the peg. 
% Block pushing has greater stability in measuring reaction force since there is only one point of contact between the pusher and block. 

\begin{table}[tb]
\vspace{4pt}
\caption{Goal cost parameters for each environment.}
\label{tab:cost parameters}
\centering

\begin{tabular}{|l|c|c|c|}
\hline
Term & block & peg & real peg \\ \hline
$Q$ & $\diag{[10,10,0,0,0]}$ & $\diag{[1,1,0,0,0]}$ & $\diag{[1,1,0,0,0]}$\\
$R$ & $\diag{[0.1,0.1,0.1]}$ & $\diag{[1,1]}$ & $\diag{[1,1]}$\\
\hline
\end{tabular}
\end{table}

% \section{Algorithm details}
\begin{algorithm}[tb]
\DontPrintSemicolon
\SetKwInOut{Given}{Given}
  \SetKwFunction{FE}{EnteringTrap}
  \SetKwProg{Fn}{Function}{:}{}
\Given{
$t_0$ time since end of last recovery or start of local dynamics, whichever is more recent,
$\x_{t_0}, ..., \x_t$,
$\xvelnom$,
$\ndynamics$,
$\trapthreshold$
}

        \For{$a \leftarrow t_0$ \KwTo $t - \ndynamics$}{
            \uIf{$\xdist(\x_t, \x_{a}) / (t - a) < \trapthreshold \xvelnom$}{
                \Return True\;
            }
        }
        \Return False\;
\caption{\texttt{EnteringTrap}}
\label{alg:enter trap}
\vspace{-4pt}
\end{algorithm}

% \begin{algorithm}
% \DontPrintSemicolon
% \SetKwInOut{Given}{Given}
%   \SetKwProg{Fn}{Function}{:}{}
% \Given{
% $\x_0, ..., \x_t$ since end of last recovery\\
% $\us_0, ..., \us_{t-1}$ since end of last recovery\\
% $\hat{\x}_1, ..., \hat{\x}_t$ one step predictions from $\approxnewdynamics$\\
% $\trapours$ trap set
% }
%         $\tau \leftarrow \min_{a} \xdist(\x_a, \x_{a+1}) / \xdist(\x_a, \hat{\x}_{a+1})$ \;
%         $\trapours \leftarrow \trapours \cup \{(\x_{\tau}, \us_{\tau})\}$
% \caption{\texttt{AddToTrapSet}: Select transitions to add to the trap set}
% \label{alg:add to trapset}
% \end{algorithm}

\begin{algorithm}
\DontPrintSemicolon
\SetKwInOut{Given}{Given}
  \SetKwFunction{FE}{EnteringTrap}
  \SetKwProg{Fn}{Function}{:}{}
\Given{
$\x_0, ..., \x_t$ since start recovery,
$\xvelnom$, parameters from Tab.\ref{tab:high level controller parameters}
}
    \uIf{$t < \ndynamics$}{
        \Return False\;
    }
    \uElseIf{$t > \nrecoverymax$}{
        \Return True\;
    }
    converged $\leftarrow \xdist(\x_t, \x_{t-\ndynamics}) / \ndynamics < \convergedthreshold \xvelnom$ \;
    away $\leftarrow \xdist(\x_t, \x_0) > \movedthreshold \xvelnom$\;
    \Return converged \textbf{and} away\;
\caption{\texttt{Recovered}}
\label{alg:leave recovery}
\vspace{-4pt}
\end{algorithm}

\begin{algorithm}
\DontPrintSemicolon
\SetKwInOut{Given}{Given}
\Given{\text{cost}, 
\text{model},
$\x$, 
$\Unom$, Tab.~\ref{tab:mppi parameters} parameters
}

$\perturbation \leftarrow$ $\normal(\mu, \Sigma)$ \tcp{\us{} perturbation for $\horizon$ steps}

$\U, \perturbation \leftarrow$ clip $\Unom$ to control bounds 

$\X_0 \leftarrow \x$ 

\HiLi $\cost \leftarrow 0$ \tcp{$\nrollouts \times \samples$}

\HiLi \For{$\rollout \leftarrow 0$ \KwTo $\nrollouts-1$}{
\For{$t \leftarrow 0$ \KwTo $\horizon-1$}{
$\X_{t+1} \leftarrow \text{model}(\X_t, \U_t)$ \tcp{sample rollout}\label{alg:rollout}

\HiLi $\cost_{\rollout,t} \leftarrow \text{cost}(\X_{t+1}, \U_t)$\;
\HiLi $\cost_\rollout \leftarrow \cost_\rollout + \cost_{\rollout,t}$
}
}

\HiLi $\cost \leftarrow $ mean $\cost$ across $\nrollouts$

$\U \leftarrow$ softmax mix perturbations\;
\Return $\U$

\caption{MPC Implementation: multi-rollout MPPI. Differences from MPPI \cite{MPPI} are highlighted.}\label{alg:mppi}
\end{algorithm}

\section{Environment details}
\label{ap:env dimensions}
The planar pusher is a cylinder with radius 0.02m to push a square with side length $\sidelength=0.3$m. We have $\yaw \in [-\pi,\pi]$,
$\pushalong \in [-a/2,a/2]$, $\pushmag \in [0,a/8]$, and $\pushdir \in [-\pi/4, \pi/4]$.
All distances are in meters and all angles are in radians. The state distance function is the 2-norm of the pose, where yaw is normalized by the block's radius of gyration. $\xdist(\x_1, \x_2) = \sqrt{(x_1-x_2)^2+(y_1-y_2)^2+\sqrt{a^2/6}(\yaw_1-\yaw_2)^2}$.
The sim peg-in-hole peg is square with side length 0.03m, and control is limited to $\pushmag_x, \pushmag_y \in [0, 0.03]$. These values are internally normalized so MPPI outputs control in the range $[-1,1]$.

\section{Representation learning \& GP}
\label{ap:learning details}
Each of the transforms is represented by 2 hidden layer multilayer perceptrons (MLP) activated by LeakyReLU and implemented in PyTorch. They each have (16, 32) hidden units except for the simple dynamics $\approxlatentdynamics$ which has (16, 16) hidden units, which is replaced with (32, 32) for fine-tuning. The feedforward baseline has (16, 32, 32, 32, 16, 32) hidden units to have comparable capacity. We optimize for 3000 epochs using Adam with default settings (learning rate 0.001), $\weightreconstruct = \weightmatch = 1$, and $\rexparam = 1$. For training with V-REx, we use a batch size of 2048, and a batch size of 500 otherwise.
% \section{Gaussian process implementation}
We use the GP implementation of gpytorch with an RBF kernel, zero mean, and independent output dimensions.  %autoregressive radial basis kernel so that each output dimension is independent. 
For the GP, on every transition, we retrain for 15 iterations on the last 50 transitions since entering non-nominal dynamics to only fit non-nominal data.

% \section*{Acknowledgment}

\ifCLASSOPTIONcaptionsoff
  \newpage
\fi



%\bibliographystyle{IEEEtran}

\bibliographystyle{IEEEtran}
\bibliography{myrefs}


%\vfill


\end{document}